\documentclass[letterpaper, 10 pt, conference]{ieeeconf}  

\IEEEoverridecommandlockouts                              

\usepackage{amsmath} 
\usepackage{amssymb}  
\usepackage{graphicx, booktabs, enumerate, url, pifont, textcomp, cite, xcolor}
\usepackage{multirow}
\usepackage{nth}
\usepackage{subcaption}
\usepackage{hyperref}
\usepackage{xspace}
\usepackage[capitalise]{cleveref}
\usepackage{siunitx}
\usepackage[ruled,vlined]{algorithm2e}
\usepackage{overpic}

\makeatletter
\DeclareRobustCommand\onedot{\futurelet\@let@token\@onedot}
\def\@onedot{\ifx\@let@token.\else.\null\fi\xspace}

\def\eg{\emph{e.g}\onedot} 
\def\ie{\emph{i.e}\onedot} 
 
\def\etc{\emph{etc}\onedot} 
 
\def\etal{\emph{et al}\onedot}

\title{\LARGE \bf
Fast Event-based Double Integral for Real-time Robotics
}

\author{Shijie Lin$^{1,5}$, Yingqiang Zhang$^1$, Dongyue Huang$^{2,5}$, Bin Zhou$^{3,5}$, Xiaowei Luo$^4$ and Jia Pan$^{1,\dagger}$
\thanks{$^1$S. Lin, Y. Zhang, and J. Pan are with the Department of Computer Science, The University of Hong Kong, Hong Kong SAR, China. 
{\tt \{lsj2048,zyq507,panj\}@connect.hku.hk}\endgraf
$^2$D. Huang is with The Department of Mechanical and Automation Engineering, The Chinese University of Hong Kong, Hong Kong SAR, China.
{\tt dyhuang@mae.cuhk.edu.hk}\endgraf
$^3$B. Zhou is with the School of Computer Science and Engineering, Beihang University, Beijing, China.
{\tt zhoubin@buaa.edu.cn}\endgraf
$^4$X. Luo is with the Department of Architecture and Civil Engineering, City University of Hong Kong, Hong Kong SAR, China.
{\tt xiaowluo@cityu.edu.hk}\endgraf
$^5$ Peng Cheng Laboratory, Shenzhen, Guangdong, China.\endgraf
$^\dagger$ Corresponding author.
}%
}

\begin{document}
\maketitle
\thispagestyle{empty}
\pagestyle{empty}

\begin{abstract}
        Motion deblurring is a critical ill-posed problem that is important in many vision-based robotics applications. The recently proposed event-based double integral (EDI) provides a theoretical framework for solving the deblurring problem with the event camera and generating clear images at high frame-rate.
        However, the original EDI is mainly designed for offline computation and does not support real-time requirement in many robotics applications. 
        In this paper, we propose the fast EDI, an efficient implementation of EDI that can achieve real-time online computation on single-core CPU devices, which is common for physical robotic platforms used in practice. 
        In experiments, our method can handle event rates at as high as 13 million event per second in a wide variety of challenging lighting conditions. We demonstrate the benefit on multiple downstream real-time applications, including localization, visual tag detection, and feature matching. 
\end{abstract}

\section{Introduction}
Clear image acquisition is the prerequisite of various vision-based algorithms to operate in robotics systems. However, due to the inherent blurry effect of the active pixel sensor, motion artifacts can heavily degrade the image when relative motion exists between the camera and scenes. This effect is further intensified in low light conditions due to the demand for longer exposure time. Condisering only frames, motion deblurring is an ill-posed problem \cite{purohit2019bringing} and poorly addressed by existing methods \cite{sun2015learning,tao2018scale,pan2018depth}. Luckily, a novel type of bio-inspired neuromorphic vision sensor, \ie, the dynamic vision sensor (DVS) \cite{dvs128,DAVIS}, can encode light variation in high temporal resolution events, allowing the motion deblurring being effectively addressed~\cite{EDI,zhang2022unifying,scheerlinck2020fast,wang2020event,xu2021motion}. 

Pan \etal \cite{EDI} proposed the event-based double integral (EDI) model that bridges the formation of events and images, showing the effective capability in deblurring and high-rate video reconstruction. EDI was later adopted by various works \cite{xu2020eventcap,gu2021learn,wang2020eventsr} for offline processing. However, for real-time robotics, existing EDI implementations are too slow (\SI{1.5}{s} per image \cite{EDI}). To employ the promising deblurring capability brought by EDI for various robotics applications, we need to run it fast, making millions of events to be processed online without jamming the whole system. However, this is still quite challenging for the following reasons. 
\begin{itemize}
    \item \textbf{High Computational Complexity.} Vanilla EDI \cite{EDI,edipami} is formulated using latent images. Without modification, the computation complexity is coupled with the image size and requires multiple redundant frame-wise operations, significantly limiting its efficiency. 
    \item \textbf{Infeasible to Real-time.}
    When running EDI for real-time robotic systems, millions of events must be processed within frame intervals to ensure continuous captured data won't jam the whole system. However, the original implementation of EDI was targeted for offline processing and its code is difficult to be adapted for online and onboard robotic system. 
    \item  \textbf{Unknown Contrast Parameters.} To deal with the contrast parameter, related works using EDI for image deblurring usually involve time-consuming optimization \cite{EDI,wang2020joint} or learning-based inference \cite{lin2020ledvdi,zhang2022unifying}, both of which are highly time-consuming and infeasible for real-time processing with CPU-only devices.
\end{itemize}

In this work, we propose fast EDI, an efficient reformulated version of EDI that achieves real-time deblurring with an event rate up to 13 Million Ev/s in a single core CPU. To achieve this performance, we propose to evenly distribute the workload for online processing and attain the goal by reformulating the EDI model and implementing it using a novel list-based container. We further introduce a method to estimate the contrast from the hardware parameters allowing robust and efficient parameter determination. In experiments, we compare our methods with existing approaches in terms of perceptual quality and runtime, showing state-of-the-art performance. Our fast EDI allows real-time deblurring and benefits various downstream applications, including visual tag detection in highly dynamic scenes, SLAM and feature matching in low light conditions. We fully release the real-time processing code and dataset with detailed hardware parameters.

\noindent{\textbf{Contribution}}:
We propose the fast EDI, an efficient way of implementing the event-based double integral (EDI), unlocking the high-speed sensing capability of event cameras for clear image acquisition. We demonstrate that the EDI can be implemented in real-time using a single core CPU, enabling the use of various applications like localization, mapping, tracking, \etc. We release an efficient C++ implementation and dataset with hardware settings. \hyperlink{https://github.com/eleboss/fast_EDI}{Webpage: github.com/eleboss/fast\_EDI}

\section{Related Work}
\noindent{\textbf{Motion deblurring and reconstruction}}.
The problems of motion deblurring and image reconstruction have been widely researched over decades. Early approaches are mostly based on the assumption of the static scene and leverages gradient \cite{fergus2006removing,krishnan2011blind,sun2013edge} or non-gradient priors \cite{lai2015blur,pan2017deblurring,yan2017image}. Recent works improved the performance through learning-based methods \cite{sun2015learning,tao2018scale,pan2018depth}. But their performance is generally not guaranteed under all conditions. Challenging motions or non-informative scenes could lead to degraded performance. With events, the motion deblurring becomes much easier to achieve. Early works fuse events and images using the asynchronous complementary filter~\cite{scheerlinck2018continuous}, manifold regularisation~\cite{munda2018real}, or direct integration~\cite{6865228}. But as the image formation model is not considered, their performance is generally worse than EDI \cite{EDI}, which takes into account the event-frame generation model and shows promising performance in image deblurring and high-rate video reconstruction. Based on EDI, later works improved their performance using learning frameworks. Songnan \etal \cite{lin2020learning} leveraged the discrete EDI as a physical insight when designing the learning framework, which was later improved by Xu \etal \cite{xu2021motion} by employing the optical flow of the event and frame for learning and training on the real data. Recently, Xiang and Yu \cite{zhang2022unifying} leveraged the EDI model to design a learnable double integral network, which can provide a generalizable model for deblurring and frame interpolation. However, previous works that rely on EDI require time-consuming optimization or learning-based inference, making real-time processing infeasible for robots with only CPU mounted. Even worse, according to our knowledge, there is no solution existing for efficient online processing of EDI. These issues significantly limit the usage of EDI in real-time robotics like localization, feature tracking, visual tag detection, \etc. 

\noindent{\textbf{EDI related applications}}. As an important model for event-frame relation, EDI has been adopted in various applications, including image denoising~\cite{wang2020joint}, high dynamic range imaging~\cite{wang2021asynchronous}, event-based super-resolution~\cite{wang2020eventsr}, human-pose estimation~\cite{xu2020eventcap}, event-based optical flow estimation~\cite{pan2020single}, and depth estimation~\cite{gehrig2021combining}. However, existing works primarily work in an offline manner that does not require real-time processing. Whenever the number of events for processing within each second easily exceeds millions, EDI computation could consume tremendous time and bottleneck the whole system. Our fast EDI can overcome this limitation and allow the use of the EDI in a wide variety of applications.
\begin{figure}[t]
        \centering
        \includegraphics[width=\columnwidth]{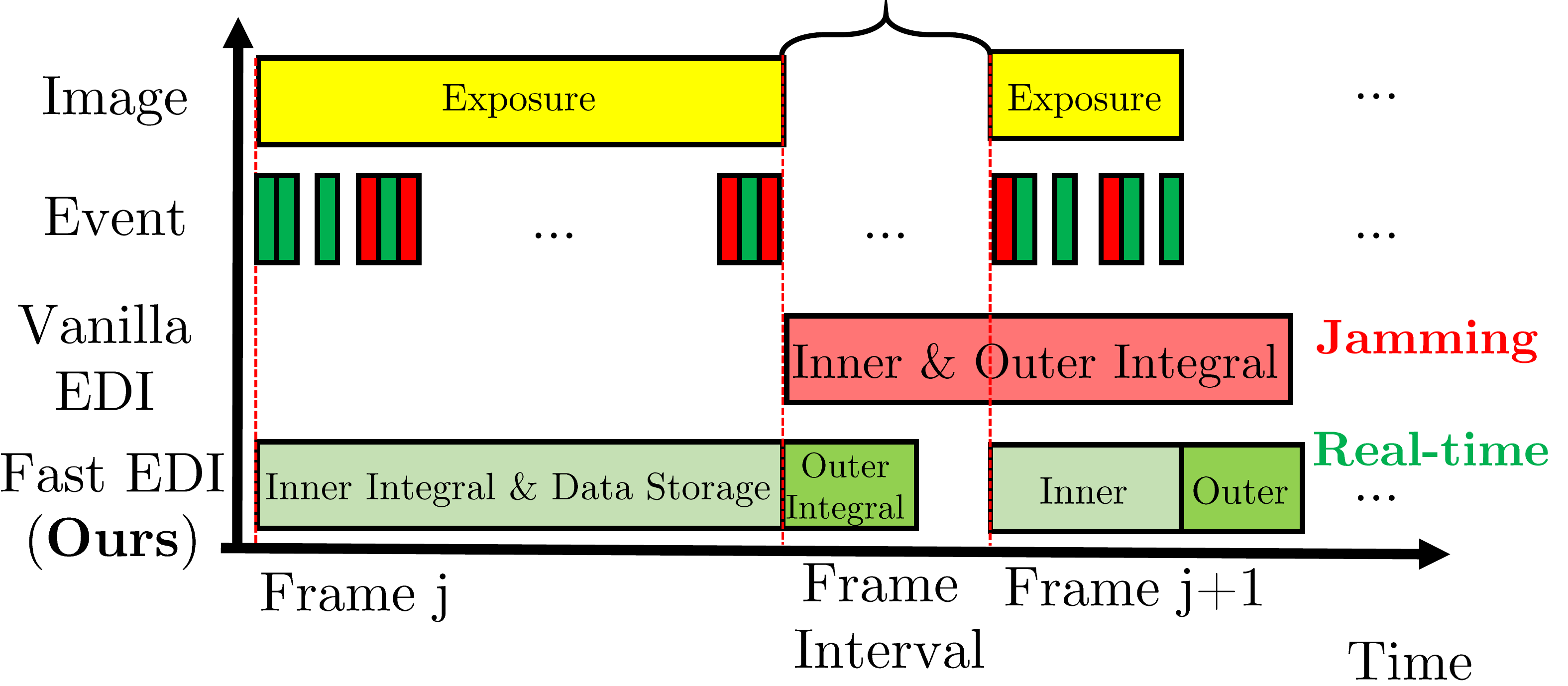}%
         \caption{Workload distribution. Our fast event-based double integral (EDI) can distribute the overall workload during acquisition time, achieving real-time processing. Vanilla EDI processes all events after the acquisition, which cannot complete before the next image is ready, leading to system jamming.}
         \label{fig:timing}
          \vspace{0cm}
\end{figure}
\section{Preliminaries}
We first briefly review the formation model of events and images, and then formulate the event-based double integral (EDI) model. 

\subsubsection*{Event generation}
The event camera works asynchronously in responding to changes of log intensity and triggers timestamped events whenever the log-scale intensity change exceeds the contrast parameter $c>0$, \ie,
$\log\left(L(\mathbf{x}, t)\right) - \log\left(L(\mathbf{x}, \tau)\right) = p \cdot c,$ where $L(\mathbf{x}, t)$ and $L(\mathbf{x}, \tau)$ indicate the instantaneous latent image at time $t$ and $\tau$ at pixel position $\mathbf{x}$, and the polarity $p\in\{+1,-1\}$ denotes the direction of intensity changes. Thus the event is a 4-dimensional tuple $\mathbf{e}_{k} \triangleq  (\mathbf{x}_k, t_k, p_k)$, where $\mathbf{x}_k = (x_k, y_k)^T$ is the pixel position, $t_k$ is the triggering time and $p_k$ is the polarity of $k$-th events, respectively. Finally, the brightness motion can be encode in the events set $\mathcal{E}=\left\{\mathbf{e}_{k} \mid k=1, \ldots, N_{ev}\right\}$, where $N_{ev}$ is the number of events. Leveraging events, we can give the relationship between latent images $L(f)$ \cite{EDI} (omitting the pixel positions $\mathbf{x}$ for readability): 
\begin{align}
L(t) = L(f)\exp\left(c\int_f^t e(s) ds \right),
\end{align}
where $e(t)\doteq p \cdot \delta(t-\tau)$ is the continuous sampling function of events with the Dirac function $\delta(\cdot)$.

\subsubsection*{EDI model}
On the other hand, the blurry images are formulated using latent images $L(t)$ within the exposure time $T$: 
\begin{align}
B = \frac{1}{T} \int_{t\in\mathcal{T}} L(t) dt,
\end{align}
where the blurry image is averaged results of latent images with the exposure duration $\mathcal{T}$. Then  the latent images is reformulated as \cite{EDI}: $L(f) = \frac{B}{E(f,\mathcal{T})}$, with 
\begin{align}
        &E(f,\mathcal{T}) = \frac{1}{T} \int_{t\in\mathcal{T}} \exp{\left(c\int_f^t e(s)ds \right) dt} \label{eq:edi}
\end{align}
indicating the physical relation between latent images and blurry images from concurrent events, which is also refer to the event-based double integral (EDI) model. 

\begin{figure}[t]
        \centering
        \includegraphics[width=0.8\columnwidth]{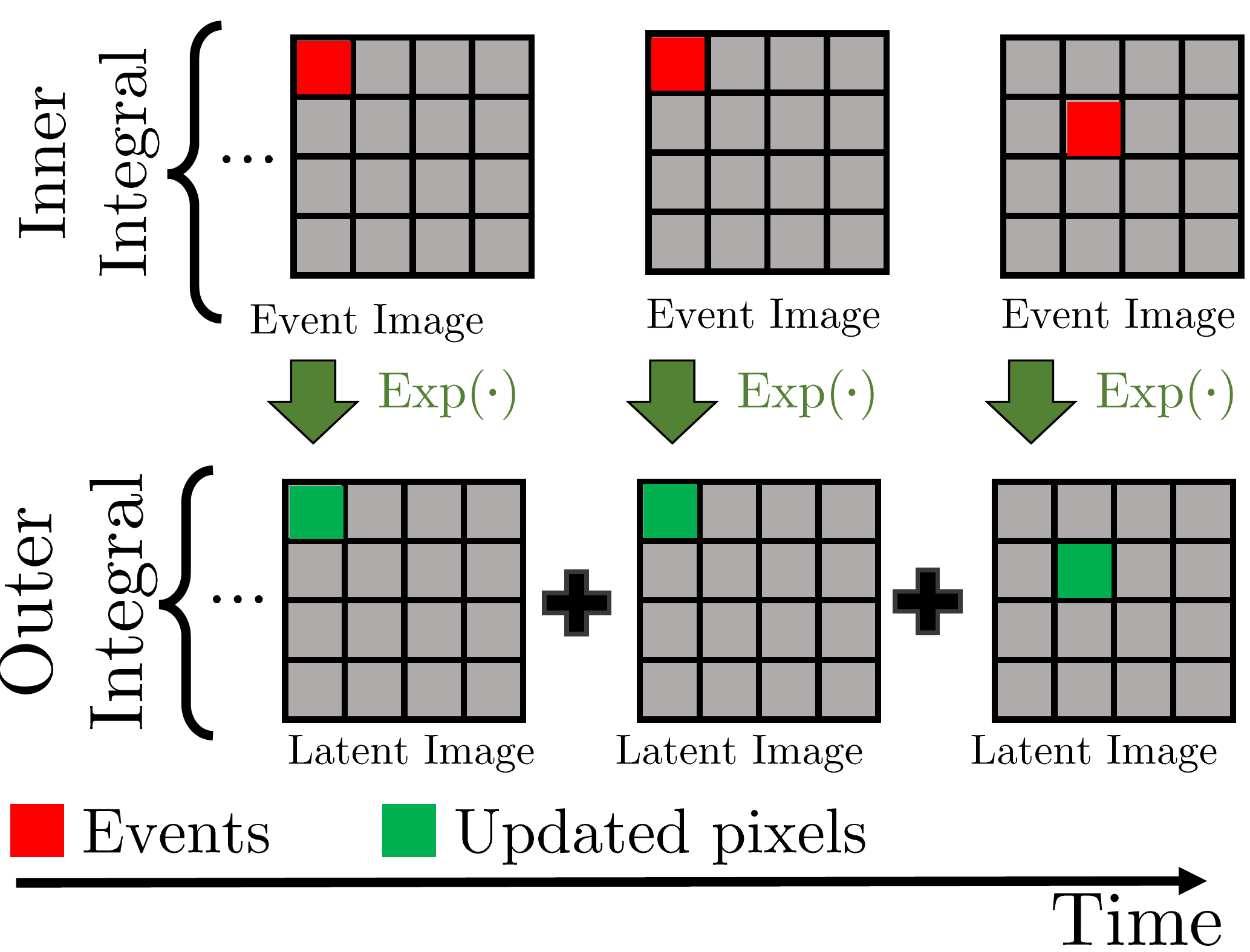}%
         \caption{EDI implemented with conventional array-like container requires frame-wise additions for the outer integral and thus is inefficient.}
         \label{fig:array-like}
\end{figure}

\begin{figure*}[t]
    \centering
  \begin{subfigure}{0.9\linewidth}
          \centering
        \includegraphics[clip,width=\columnwidth]{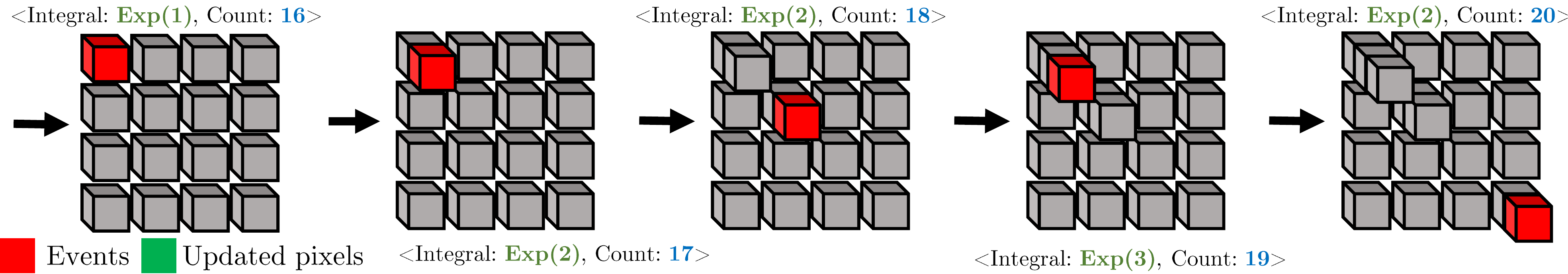}%
        \caption{Step 1: inner integral (during exposure)}
  \end{subfigure}
  \hfill
  \begin{subfigure}{0.9\linewidth}
          \centering
        \includegraphics[clip,width=\columnwidth]{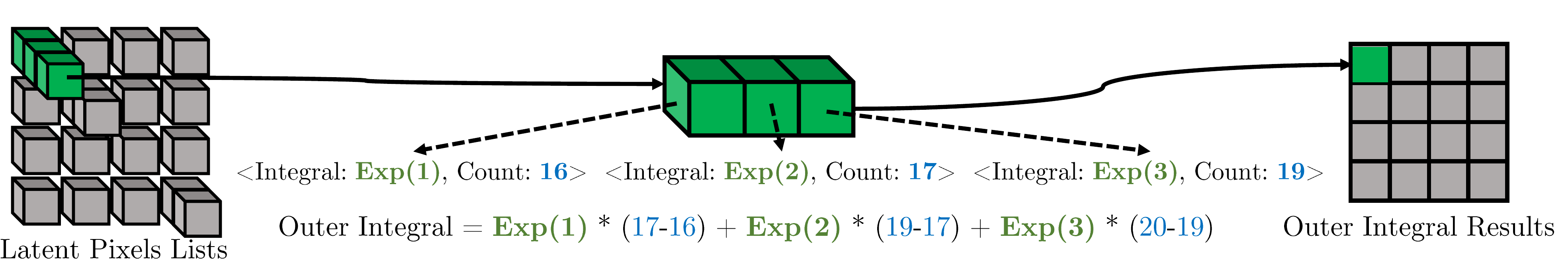}%
        \caption{Step 2: outer integral (after exposure).}
  \end{subfigure}
   \caption{EDI implemented with the proposed list-based container can significantly reduce the number of addition operations and decouple it from the image size, thus achieving efficient real-time processing. In step 1, the inner integral operates pixel-wise and pushes back the latest latent pixel value, \ie, $\text{Exp}(\cdot)$ and the counter value in the latent pixels lists. In step 2, the outer integral could be computed by summing the multiplication between latent pixel value (green exp value) and the difference of two consecutive counter numbers.}
   \label{fig:list-based}
 \vspace{0cm}
\end{figure*}

\subsubsection*{Current issues}
Although the continuous EDI model is given in previous literature \cite{EDI,edipami}, efficient online implementation was never specified. To achieve this goal, we need to overcome the following difficulties: 
\begin{itemize}
    \item \textbf{Workload imbalance.} The EDI computation requires an accurate selection to separate events within the exposure interval. However, for online processing, the camera generates data sequentially (\cref{fig:timing}), meaning the selection cannot be made before the image's exposure is completed. Thus, vanilla EDI starts processing after the exposure is completed. However, the exposure time of an image could exceed \SI{30}{ms}, and millions of events could be triggered during this time. In a real-time system, all data need to be processed before subsequent data is ready. If the computation of vanilla EDI cannot be done within the frame interval, the subsequent data will cause jamming, slowing down the system and leading to an out-of-memory crash. 
    \item \textbf{Redundant operations.}  Unfortunately, the vanilla EDI usually cannot complete all computations within the frame interval due to redundant operations. Specifically, computing \cref{eq:edi} requires repeatedly frame-like reconstruction of the latent image (inner integral) over the whole exposure interval (outer integral), making the real-time processing highly challenging. 
    \item \textbf{Unknown contrast parameter.} Previous works \cite{EDI,edipami} use time-consuming optimization or inaccurate manual setting to handle the contrast parameters $c$. Both of them cannot satisfy the need for real-time robotics systems. 
\end{itemize}

Therefore, we need to develop the whole computation pipeline from scratch to implement the EDI for efficient online processing.




\section{Methods}
To improve the EDI for real-time processing, we developed the fast EDI, 
whose is implementation is based on a novel list-based container to achieve a balanced workload distribution. We further introduce the method of contrast estimation using hardware parameter \cite{nozaki2017temperature}, allowing a robust and efficient determination.

\subsection{Fast event-based double integral (F-EDI)}
\subsubsection{Workload distribution} 
As the events are generated asynchronously from the sensor, one key to boosting the processing speed is to make each event processed right after it is triggered. However, for robotics systems, various applications are needed to consume the processing power of the CPU, and cores in the low-end processor are limited. Thus it is best if we can complete all computation using a single thread. To do that, we first compute the inner integral and exponentiation for each event \ie
$\exp\left(c\int_f^t e(s)ds\right)$, and record the integral result for quick retrieval. Once the exposure is completed, the outer integral only needs to retrieve all containers and calculate the outer integral. This workload distribution allows us to fully leverage the CPU's computing resource to process millions of events in real-time (\cref{fig:timing}), thereby alleviating the jamming problem. 

\subsubsection{Efficient list-based EDI} 
However, even with the desired workload, the EDI computation is still inefficient for real-time processing. 
The vanilla EDI is formulated as a latent image, and the inner integral is responsible for rendering the latent image. Then the outer integral needs to sum up all latent images repetitively. However, since exponentiation exists between the inner and outer integral, frame-wise additions are required for summing up all latent images. As shown in \cref{fig:array-like}, given the latest triggered events marked in red, the inner integral renders the event images, then exponentiation transforms the event image into the latent image, and the outer integral need to sum up all latent images. The first two operations, \ie inner integral and exponentiation, are efficient as we can use the pixel-wise operation. But after the exponentiation, the latent image is nonlinear. Thus the outer integral needs to repetitively sums up all latent images in a frame-wise manner. Meaning for each event, equal to frame size $N_{x}$ additions are required for the outer integral leading to $\mathcal{O}(N_{ev} \times N_{x})$ complexity.

We find this process could be simplified by redesigning the container. Since each time an event is generated, only one pixel alters its value in the latent image, most pixels for the additions in outer integral remain unchanged. Thus, we can temporally record each pixel's inner integral results and the number of additions during the exposure time. Then, we can convert the repetitive additions to summing several multiplications in a single retrieval after the exposure time. Therefore, we propose the list-based container, as shown in \cref{fig:list-based}. With this container, the fast EDI can be done in two steps. \textbf{Step 1:} During the exposure time, it conducts the inner integral using the latest activated events, taking its exponentiation, and increasing the global addition counter by one. Then two values (\ie, $<$exp value, counter$>$) will be pushed back in a list marked by specific pixels positions (the red cube in \cref{fig:list-based}). \textbf{Step 2:} After exposure, our fast EDI loops over all lists at specific pixels to retrieve the integral values and compute the outer integral in a each list by summing multiplications between the latent pixel value (green exp value in \cref{fig:list-based})) and the difference of two consecutive counter numbers (blue value in \cref{fig:list-based}). 

In this way, we can turn millions of additions into simply hundreds of multiplications and additions, thereby reducing the processing time. Finally, the time complexity is purely linear to the number of events, \ie, $\mathcal{O}(N_{ev})$.

\subsection{Event Contrast from Hardware Parameters}
Vanilla EDI \cite{EDI} uses an energy function based on image sharpness for optimization to find the optimal contrast. But such an optimization algorithm has two drawbacks. First, if the highly blurred image lacks enough high-contrast edges, the optimization is likely to converge to a local optimum. Second, the optimization is inefficient, making it difficult to deploy on real-time robotic systems. To address these issues, we introduce an estimation method based on hardware parameters. From hardware settings, we can directly calculate its contrast: 
\begin{align}
    C_\text{OFF}=\frac{\kappa_{n} \mathcal{C}_{2}}{\kappa_{p}^{2} \mathcal{C}_{1}} \ln (\frac{\mathcal{I}_{\text{OFF}}}{\mathcal{I}_{d}}), 
    C_\text{ON}=\frac{\kappa_{n} \mathcal{C}_{2}}{\kappa_{p}^{2} \mathcal{C}_{1}} \ln (\frac{\mathcal{I}_{\text{ON}}}{\mathcal{I}_{d}}),   
\end{align}
where $\kappa_{n} = \kappa_{p} =  0.7$ are the back gate coefficients of n and p FET transistors. $\mathcal{C}_{1}/\mathcal{C}_{2}$ is the capacitor ratio of DVS ($130/6$ for our DAVIS346). $\mathcal{I}_{\text{d}}$, $\mathcal{I}_{\text{ON}}$, and $\mathcal{I}_{\text{OFF}}$ are the bias current set by the user in a coarse-fine bias generator~\cite{delbruck201032}, which can be computed with jAER toolbox~\cite{jaer}. In this way, we can greatly simplify the optimization and increase the efficiency.
\section{Experiments}
In experiments, we first evaluate the primary performance of the fast EDI, \ie, runtime and deblurring capability. Then we evaluate the fast EDI in three real-time downstream applications, \ie, feature tracking, visual tag detection, and SLAM. We encourage the reader to view the supplementary video for more information.
\begin{table}[t]
        \centering
        \caption{Average scores of deblurring (Best: $\downarrow$) }
        \label{tab:deblur}
        \resizebox{0.9\linewidth}{!}{
            \begin{tabular}{c|cccc}
                \hline
                \textbf{Metric} & \multicolumn{4}{c}{\textbf{Method}} \\
                       &\cite{darkchannel} &\cite{deeplearning} &\cite{EDI} &Ours \\
                \hline
                PIQE~\cite{venkatanath2015blind}    &62.92              &54.71               &42.55      &\textbf{38.12} \\
                SSEQ~\cite{liu2014no}    &31.64              &37.58               &32.04      &\textbf{24.01} \\
                BRISQUE~\cite{mittal2012no} &32.53              &37.39               &35.04      &\textbf{28.75} \\
                \hline               
        \end{tabular}
        }     
      \vspace{0cm}
\end{table}
\begin{table}[t]
   \centering
   \caption{Average matched inliers of feature tracking (Better: $\uparrow$)}
   \label{tab:inliersoffeaturematching}
   \resizebox{\linewidth}{!}{
       \begin{tabular}{c|cc|cc|cc}
           \hline
            \textbf{Feature} &  \multicolumn{2}{c|}{\textbf{Trajectory 1}}   &\multicolumn{2}{c|}{\textbf{Trajectory 2}} &\multicolumn{2}{c}{\textbf{Trajectory 3}}   \\
           
                                        & Blur     &Ours         &Blur &Ours          &Blur &Ours  \\
            \hline
            SURF~\cite{bay2008speeded}  & 83.5  &\textbf{92.1} &90.2                  &\textbf{100.4} &95.5                       &\textbf{106.8} \\
            FAST~\cite{rosten2005fusing}& 39.3  &\textbf{41.6} &42.7                  &\textbf{46.1}  &71.2                        &\textbf{73.6} \\
            MSER~\cite{matas2004robust} & 39.2  &\textbf{42.3} &48.8                  &\textbf{51.6} &82.0                       &\textbf{84.6} \\
            BRISK~\cite{2011brisk}      & 68.3  &\textbf{67.8} &73.2                  &\textbf{73.3} &112.2                       &\textbf{112.4} \\
            \hline                           
   \end{tabular}
   }     
 \vspace{0cm}
\end{table}
\begin{figure*}[t]
  \begin{subfigure}{0.16\linewidth}
        \includegraphics[clip,width=\columnwidth]{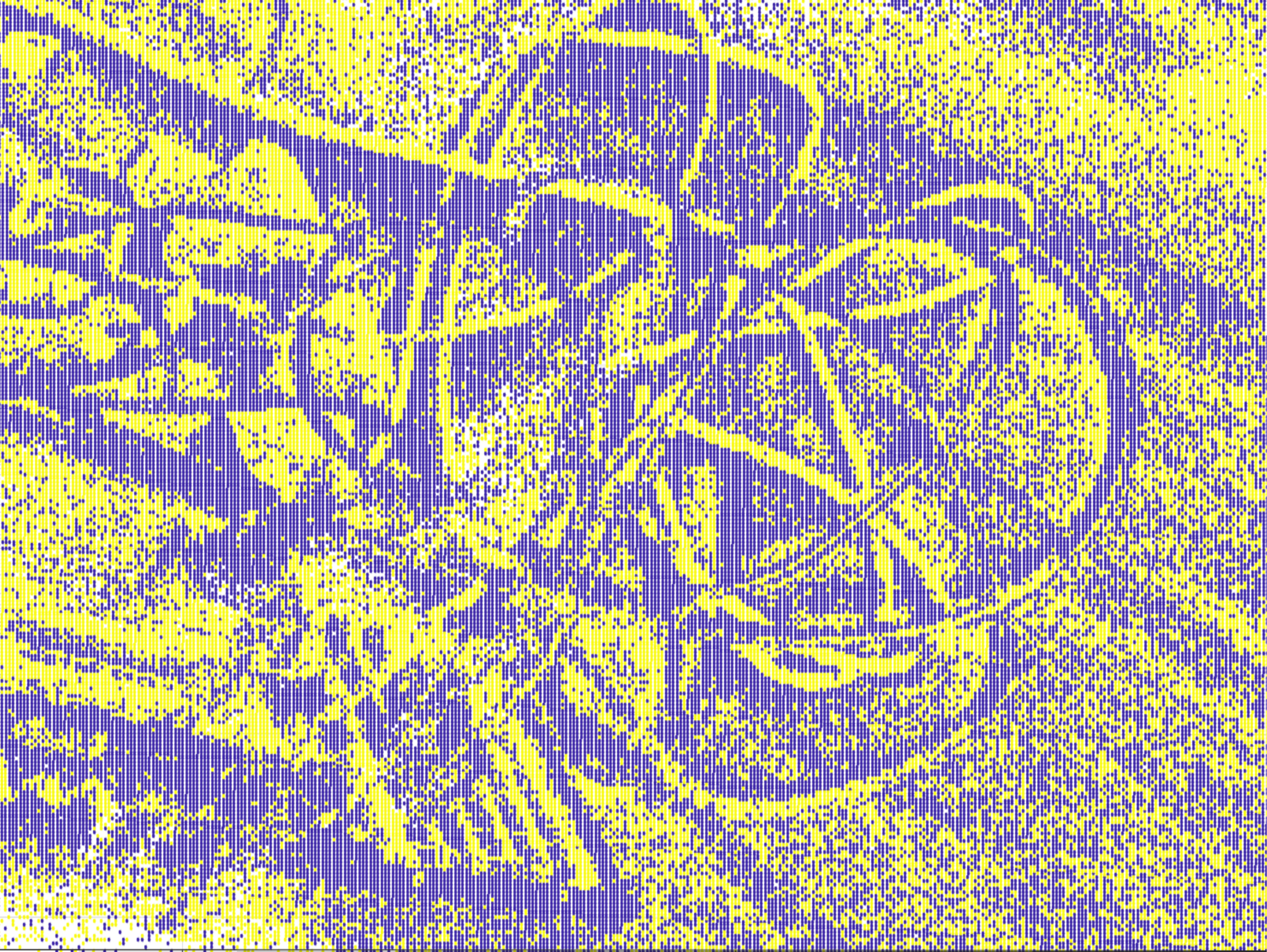}%
  \end{subfigure}
  \hfill
  \begin{subfigure}{0.16\linewidth}
        \includegraphics[clip,width=\columnwidth]{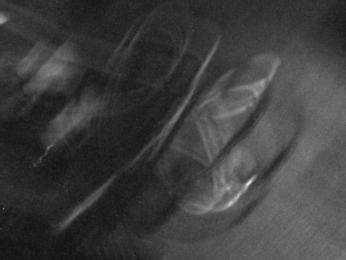}%
  \end{subfigure}
   \hfill
   \begin{subfigure}{0.16\linewidth}
        \includegraphics[clip,width=\columnwidth]{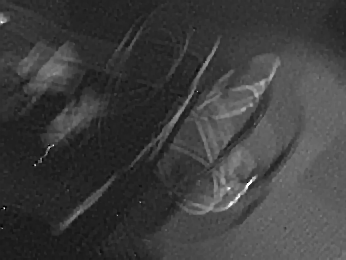}%
  \end{subfigure}
  \hfill
  \begin{subfigure}{0.16\linewidth}
        \includegraphics[clip,width=\columnwidth]{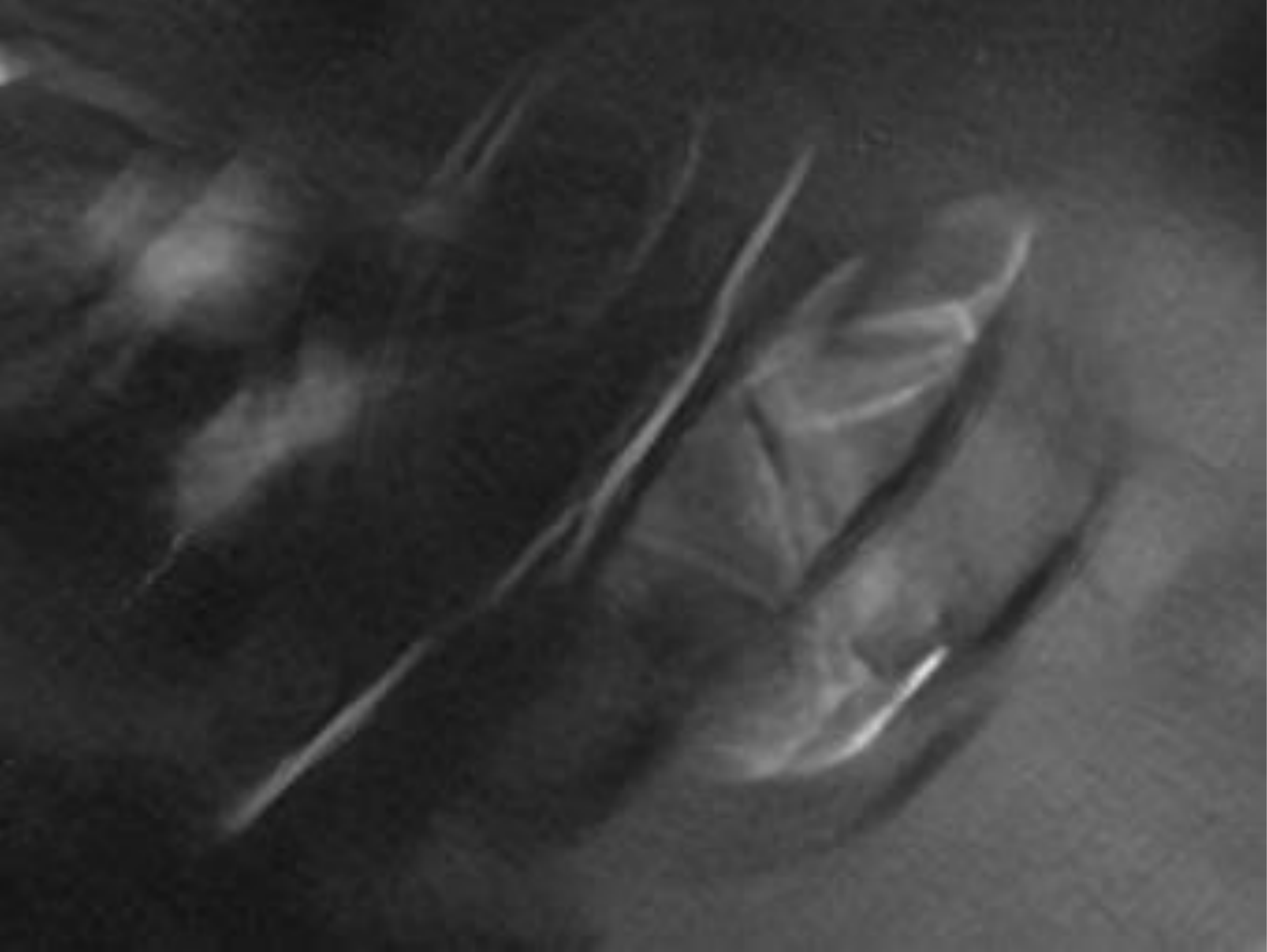}%
  \end{subfigure}
  \hfill
  \begin{subfigure}{0.16\linewidth}
        \includegraphics[clip,width=\columnwidth]{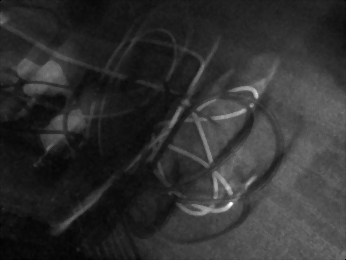}%
  \end{subfigure}
  \hfill
  \begin{subfigure}{0.16\linewidth}
        \includegraphics[clip,width=\columnwidth]{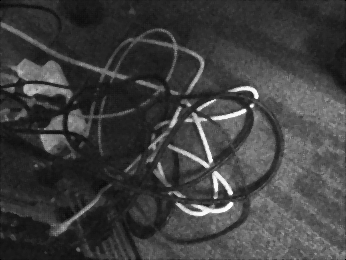}%
  \end{subfigure}
  \\
  \\
  \begin{subfigure}{0.16\linewidth}
    \includegraphics[clip,width=\columnwidth]{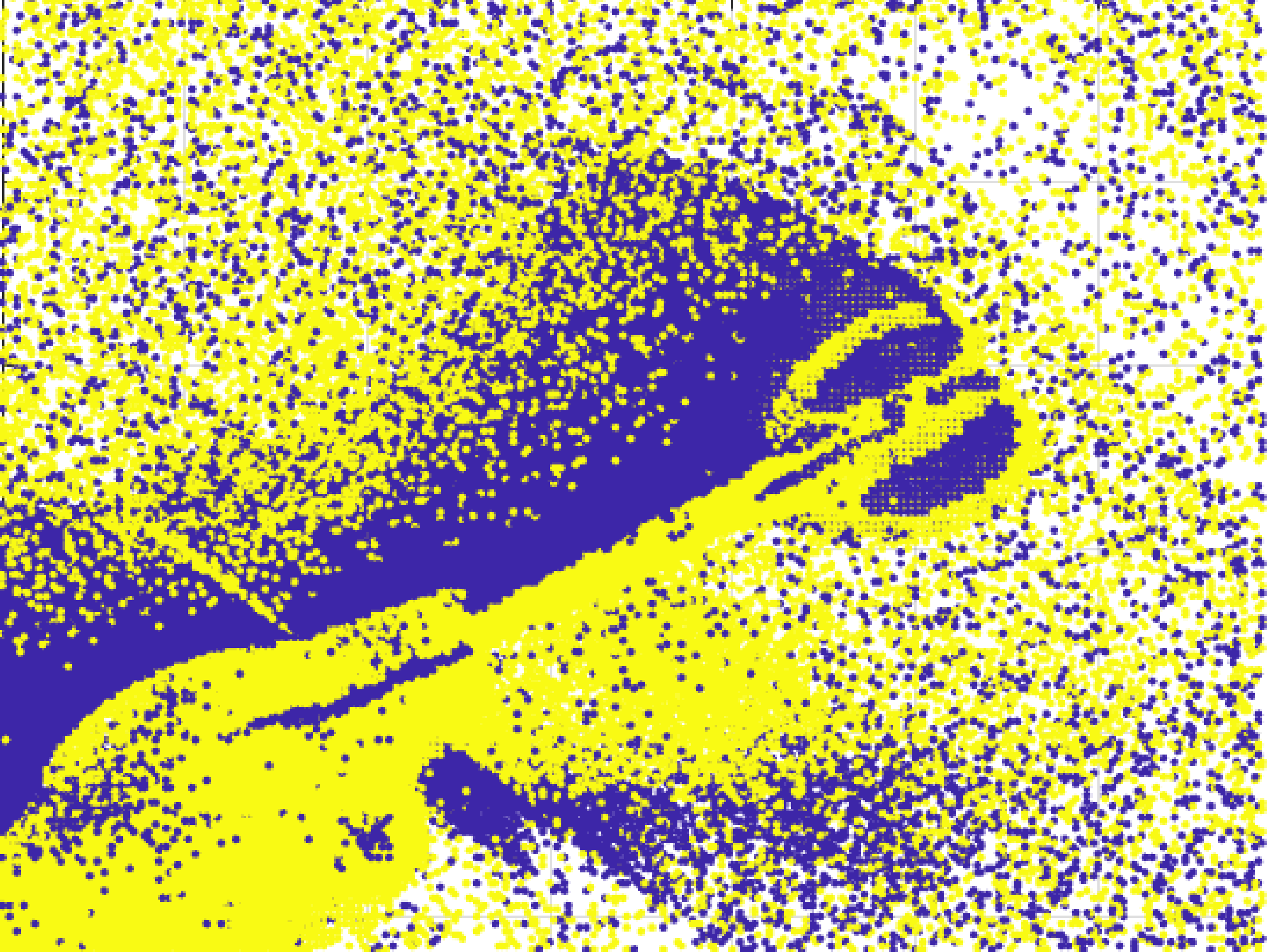}%
    \caption{Events}
  \end{subfigure}
  \hfill
  \begin{subfigure}{0.16\linewidth}
    \includegraphics[clip,width=\columnwidth]{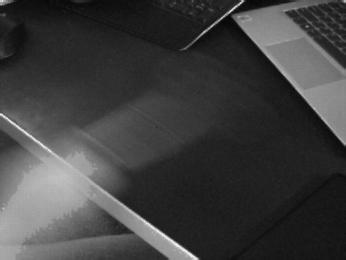}%
    \caption{Images}
    \label{fig:scissor_image}
  \end{subfigure}
  \hfill
  \begin{subfigure}{0.16\linewidth}
    \includegraphics[clip,width=\columnwidth]{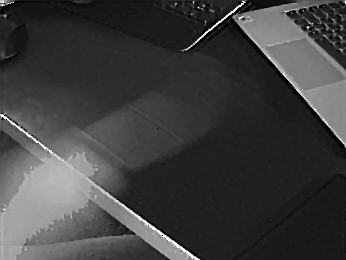}%
    \caption{\cite{darkchannel}}
  \end{subfigure}
  \hfill
  \begin{subfigure}{0.16\linewidth}
    \includegraphics[clip,width=\columnwidth]{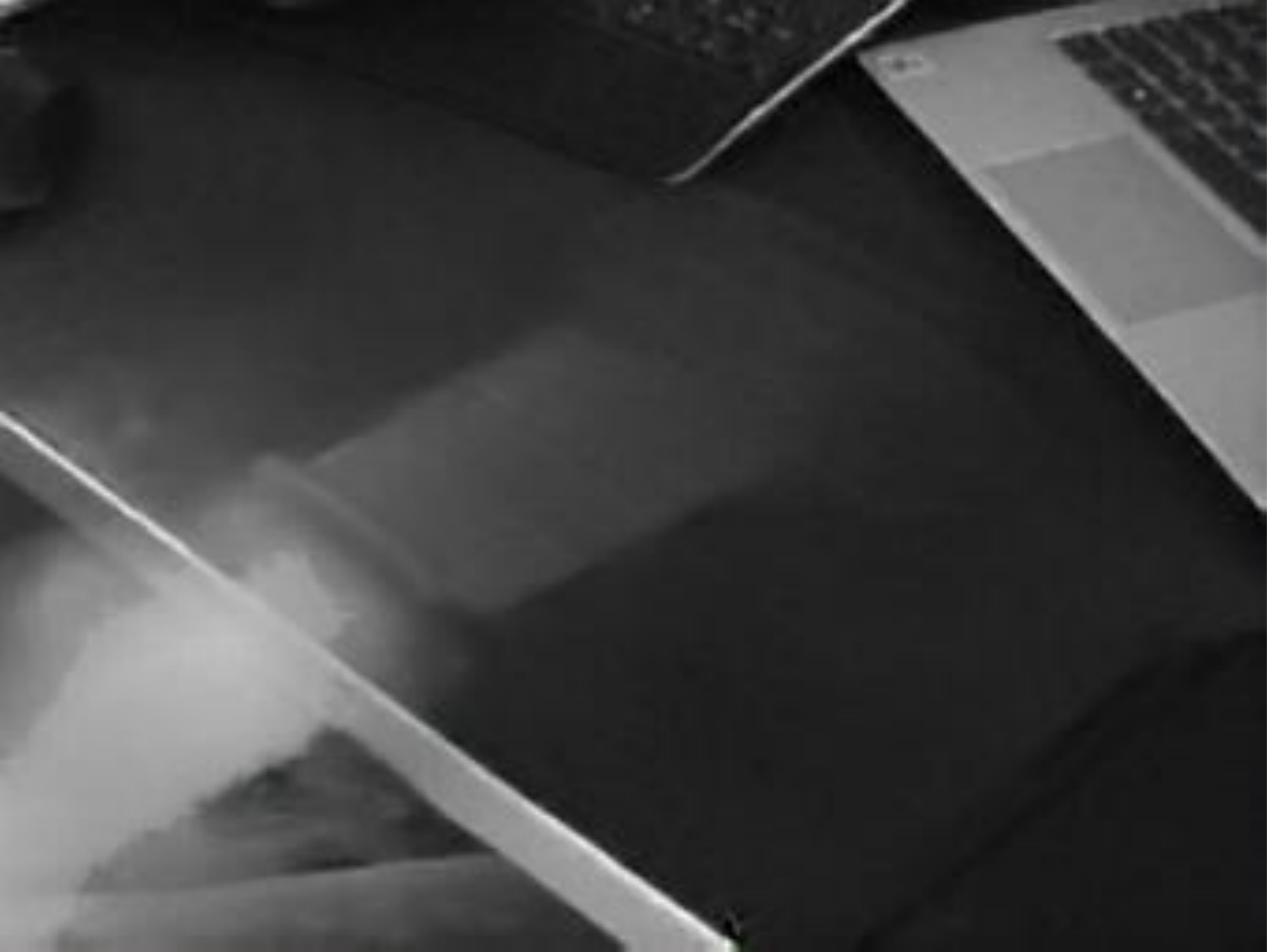}%
    \caption{\cite{deeplearning}}
  \end{subfigure}
  \hfill
  \begin{subfigure}{0.16\linewidth}
    \includegraphics[clip,width=\columnwidth]{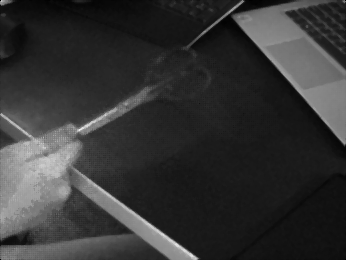}%
    \caption{\cite{EDI}}
    \label{fig:scissor_edi}
  \end{subfigure}
  \hfill
  \begin{subfigure}{0.16\linewidth}
    \includegraphics[clip,width=\columnwidth]{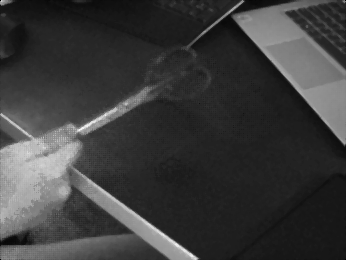}%
    \caption{Ours}
    \label{fig:scissor_our}
  \end{subfigure}
   \caption{Image deblurring results in two sequences, \textit{line} (first row) and \textit{scissor} (second row). (c) Conventional method~\cite{darkchannel} and (d) learning-based method~\cite{deeplearning} fail to recover (a) the heavily blur images. (e) \cite{EDI} works well when the background is clean but is unable to tackle heavily blurred images. In contrast, (f) our method can stably generate clean images.}
   \label{fig:deblur}
 \vspace{0cm}
\end{figure*}
\begin{figure*}[t]
        \centering
        \begin{subfigure}{0.245\linewidth}
          \includegraphics[width=\columnwidth]{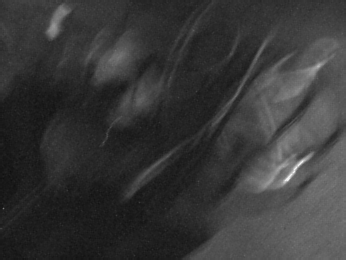}%
          \caption{Blurred image}
          \label{fig:1614177820794554_EC}
        \end{subfigure}
        \hfill
        \begin{subfigure}{0.245\linewidth}
          \includegraphics[width=\columnwidth]{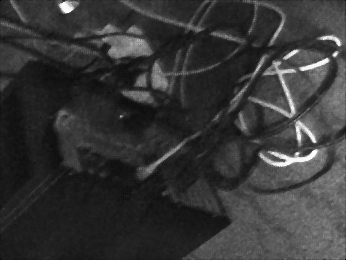}%
          \caption{Our deblurred image}
          \label{fig:1614177820794554_deblur}
        \end{subfigure}
        \hfill
        \begin{subfigure}{0.49\linewidth}
          \includegraphics[width=\columnwidth]{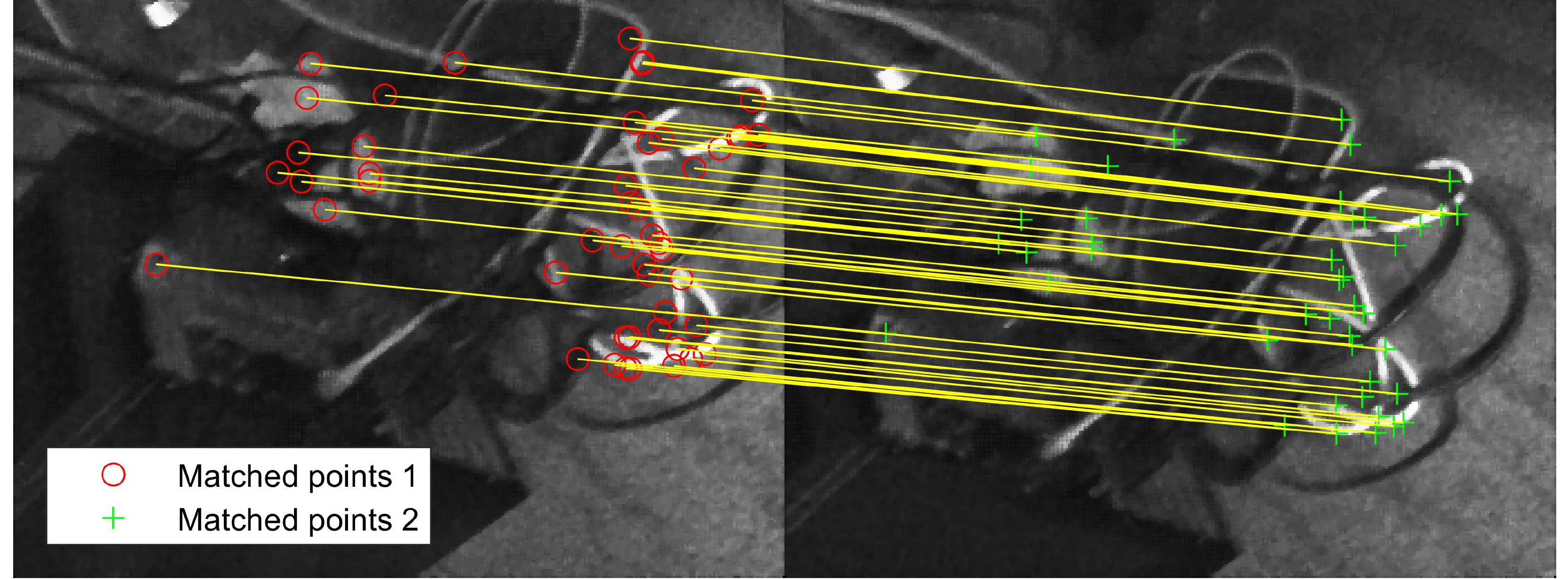}%
          \caption{Matching results enhanced by our fast EDI}
          \label{fig:feature_match}
        \end{subfigure}
        \caption{Results of SURF~\cite{bay2008speeded} feature matching. Our method can effectively deblur (a) the blurry image and generate clear images for (b) accurate feature matching.}
      \vspace{0cm}
        \label{fig:featurematching}
\end{figure*}
\begin{figure*}[t]
  \centering
  \begin{subfigure}{0.3\linewidth}
    \includegraphics[clip,width=\columnwidth]{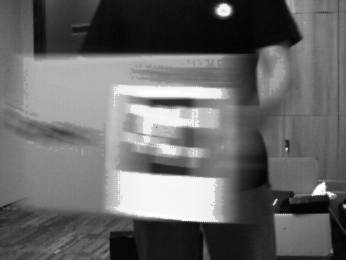}%
    \caption{Blur}
    \label{fig:apblur}
  \end{subfigure}
  \hfill
  \begin{subfigure}{0.3\linewidth}
    \includegraphics[clip,width=\columnwidth]{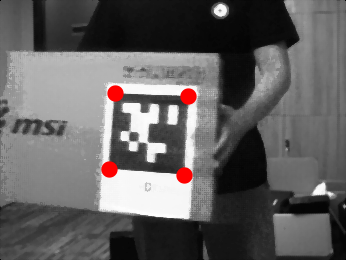}%
    \caption{Result 1}
    \label{fig:success}
  \end{subfigure}
  \hfill
  \begin{subfigure}{0.3\linewidth}
    \includegraphics[clip,width=\columnwidth]{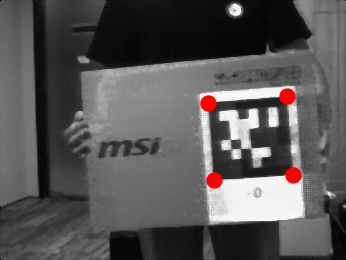}%
    \caption{Result 2}
    \label{fig:success1}
  \end{subfigure}
  \caption{Results of Apriltag~\cite{olson2011apriltag} detection. Our method can deblur (a) the blurry image cause by rapid shaking motions and generate (b) (c) clear images, making consecutive successful Apriltag detection possible.}
   \label{fig:apriltagdec}
  \vspace{0cm}
\end{figure*}
\begin{figure*}[t]
        \centering
        \begin{subfigure}{0.495\linewidth}
                \includegraphics[clip,width=0.49\columnwidth]{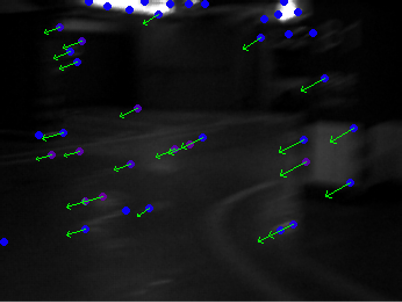}%
                \hfill
                \includegraphics[clip,width=0.49\columnwidth]{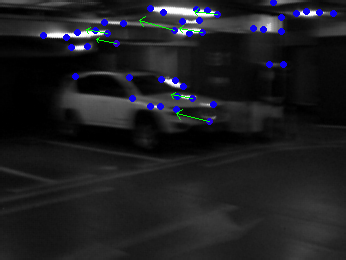}%
                \caption{Tracking results using original blur images}
                \label{fig:slam_tracking_blur}
        \end{subfigure}
        \hfill
        \begin{subfigure}{0.495\linewidth}
                \includegraphics[clip,width=0.49\columnwidth]{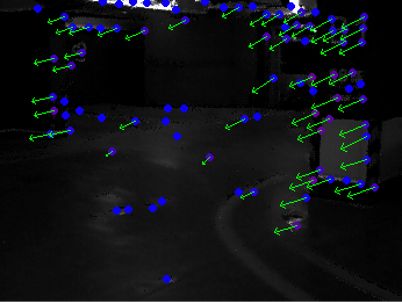}%
                \hfill
                \includegraphics[clip,width=0.49\columnwidth]{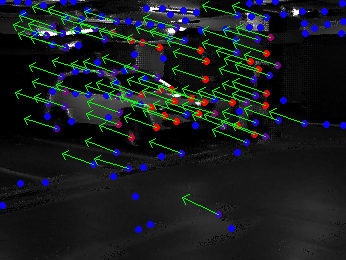}%
                \caption{Tracking results using our deblurred images}
                \label{fig:slam_tracking_deblur}
        \end{subfigure}
        \caption{Tracking result using VINS-mono SLAM~\cite{qin2017vins}. It could be clearly observed that (a) the tracking using blurry images cannot extract effective features. Most of them are in low confidence (blue dots). In contrast, tracking using images deblurred by our methods has much more effective features with high confidence (red dots).}
        \label{fig:slam}
      \vspace{0cm}
\end{figure*}

\subsection{Experimental Setups}
\subsubsection{Hardware}
We conducted all the experiments using Intel NUC mini PC with Intel i7-10710H@1.1 GHz CPU, and  DAVIS346 \cite{DAVIS} with contrast parameter set as as $C_\text{ON}=0.26$ and $C_\text{OFF}=-0.26$ to fit the single contrast model used in \cite{EDI}. The NUC mini PC is a common choice for various robotics systems and has multi-CPU cores for running other robotics algorithms like recognition, planning, and localization. We use C++ to implement our algorithms on the DV Platform with a ROS node to publish the real-time results. 
\subsubsection{Dataset} Since existing datasets do not include details of hardware settings for contrast estimation, during the real-time experiments, we recorded the online data as an evaluation dataset, which contains different motions (\eg, shaking, random move) and scene illuminations (\eg, low lighting conditions, sunlight, artificial light). Details are in the appendix.

\subsection{Runtime Performance}
The vanilla EDI is implemented using Matlab with C++ warper and cannot support further modification. Its performance also can not support real-time processing (1.5 seconds per image, according to \cite{EDI}). Thus, we compare our fast EDI with EDI implemented with the conventional array-like container using C++ and the same setup. During experiments, we record the average events rate of both methods.
Our fast EDI could exceed 13 million Ev/s event rates using a single core CPU. In comparison, vanilla EDI can only achieve 49 thousand Ev/s. This result shows our fast EDI is $260 \times$ faster than vanilla EDI, which demonstrates the efficiency of our method.

\subsection{Deblurring}
We compare our approach with other state-of-the-art image deblurring methods, including conventional method~\cite{darkchannel}, learning-based method~\cite{deeplearning}, and event-frame reconstruction method~\cite{EDI}. Qualitative comparisons are shown in \cref{fig:deblur}. Multiple no-reference metrics, \ie, SSEQ~\cite{liu2014no}, PIQE~\cite{venkatanath2015blind}, and BRISQUE~\cite{mittal2012no} are adopted to provide quantitative comparisons, with the average scores on our dataset shown in \cref{tab:deblur}, where the lower is the score, the better is the performance. According to \cref{tab:deblur}, our method achieves the best score regarding all metrics. Other methods get suboptimal scores because the heavily blurred image can hardly provide any useful information for reconstruction. From \cref{fig:deblur}, we can see that both our method and~\cite{EDI} handle the partially blurred images well, \eg, the blur of a moving scissor in \cref{fig:scissor_image}. However, for completely blurred images, only our method can reconstruct clean images (\cref{fig:scissor_our}). \cite{EDI} converges to sub-optimal contrast parameters, and the other two methods~\cite{deeplearning, darkchannel} fail to converge. It is worth noting that, with a carefully manually adjusted contrast parameter, the deblurring performance of vanilla EDI is identical to our fast EDI. Thus our method also shares the defects of EDI, like easily affected by noise. Some of these defects have been addressed in recently proposed learning-based methods~\cite{zhang2022unifying}, but such methods rely on GPU for fast inference. As our work aims at introducing an efficient implementation of EDI, comparison with them is out of the scope of this work.

\subsection{Real-time Applications}
\subsubsection{Feature matching and tracking}
Our method can also benefit the feature matching and tracking by improving image quality. As shown in \cref{fig:1614177820794554_EC}, images taken in low lighting suffer from motion blur, which can barely provide any information for feature matching. In contrast, our fast EDI recovers a clean image in \cref{fig:1614177820794554_deblur}, allowing successful SURF~\cite{bay2008speeded} feature matching in \cref{fig:feature_match}. Quantitative comparison results are summarized in \cref{tab:inliersoffeaturematching}, where our fast EDI improves the matching for all three types of features, which again verifies the effectiveness of our method. 
\subsubsection{Visual tag detection}
Our fast EDI also benefits visual tag detection due to its effectiveness in real-time image deblurring. As shown in \cref{fig:apblur}, large motions will blur the visual tag, while our reconstructed clean image can properly preserve important information for successful Apriltag~\cite{olson2011apriltag} detection in a sequence of 206 Apriltag images, as shown in \cref{fig:success} and \cref{fig:success1}. In particular, the number of detected Apriltag increases from 73 to 198 after using our method.
\subsubsection{SLAM}
Simultaneous Localization and Mapping (SLAM) is an important robotics application that relies on real-time image processing. We use the VINS-Mono (\cite{qin2017vins}, no loop closure) in a dark underground garage with a handheld DAVIS346 camera for evaluation. In experiments, we started and ended from the same position marked by visual tags. As the environment is quite dark ($<$ \SI{50}{Lux}), the frame-based camera requires a longer exposure time for imaging, introducing motion blur and destablizing the tracking front-end (\cref{fig:slam_tracking_blur}). Our fast EDI can provide real-time deblurred results to facilitate tracking (\cref{fig:slam_tracking_deblur}), gaining a much higher localization accuracy compared with trajectory using original blur image, as shown in \cref{fig:slam_traj}. Using blurry images, the translation error between starting and ending point is \SI{44.16}{meter}. This error is reduced to \SI{5.22}{meter} with the assistance of fast EDI.
\begin{figure}[t]

        \begin{subfigure}{\linewidth}
                \centering
            \includegraphics[width=0.8\columnwidth]{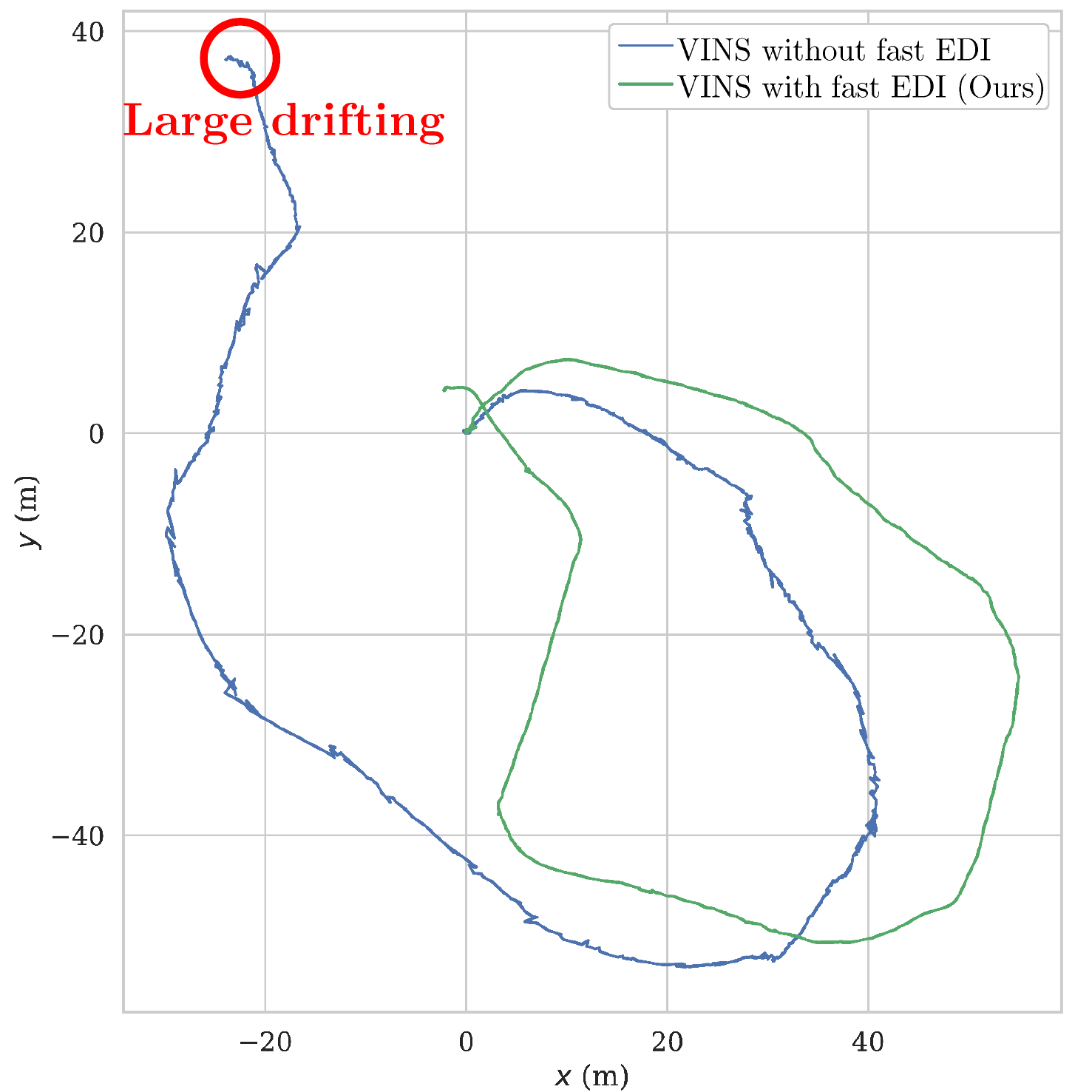}%
                \caption{Trajectories of VINS-Mono (XY-axis)}
                \label{fig:slam_traj}
                \includegraphics[width=0.8\columnwidth]{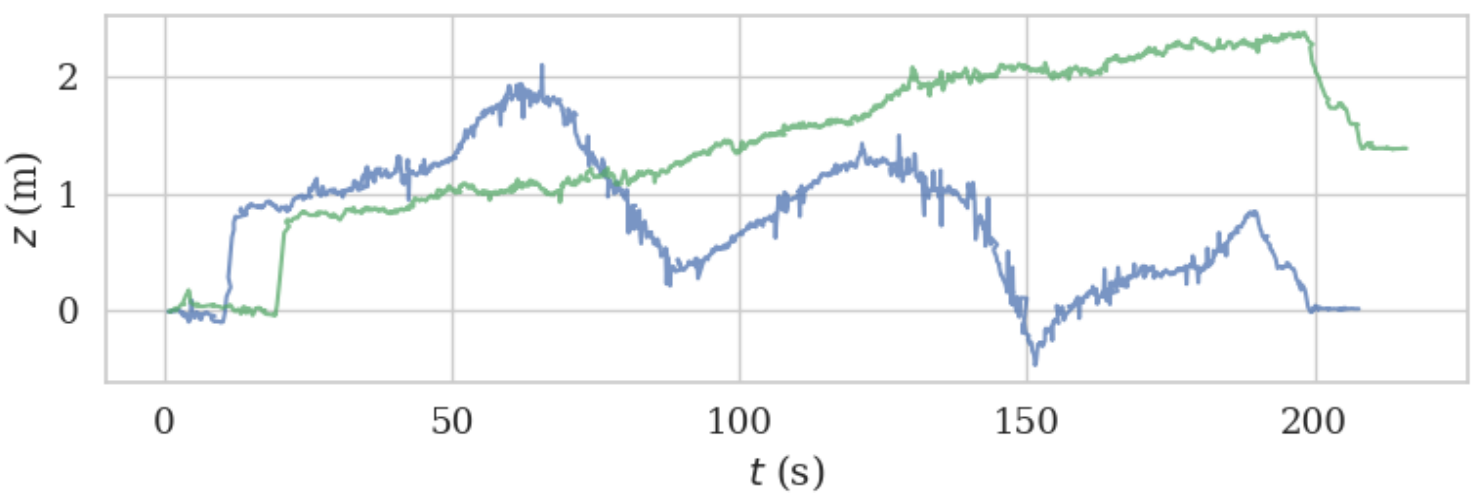}%
                \caption{Trajectories of VINS-Mono (Z-axis)}
                \label{fig:slam_traj_z}
        \end{subfigure}
        \caption{Result of SLAM using VINS-mono \cite{qin2017vins}. VINS runs with images deblurred by our fast EDI (green trajectory) has much lower drifting compared with results using original blurred image (red trajectory). The starting and ending position are the same and marked by a calibration board. Best view in color.}
        \label{fig:slam_traj}
\end{figure}

\subsection{Limitations} The memory consumption of this work increases linear to the number of triggered events. We therefore suggest that a minimum of \SI{8}{GB} memory be allocated for the system. When using event cameras from manufacturers other than Inivation, it is noteworthy that the jAER toolbox may not yield accurate contrast parameter values. Hence, it is advised that users either contact the manufacturers directly for inquiry or employ calibration methods \cite{wang2020eventcal}.

\section{Conclusion}
 In this paper, we propose the fast EDI, a novel way to efficiently implement the EDI to achieve real-time image deblurring using event cameras. In comparison with the conventional EDI model, we boost the computational speed 260 times on single core CPU and successfully bring the deblurring capability of EDI in improving the performance of various real-time applications, including SLAM, visual tag detection, and feature matching.






\bibliographystyle{IEEEtran}
\bibliography{IEEEabrv,refernence}

\end{document}